\documentclass[10pt,letterpaper]{article}
\usepackage[top=0.85in,left=2.75in,footskip=0.75in]{geometry}

\usepackage{amsmath,amssymb}

\usepackage{changepage}

\usepackage[utf8x]{inputenc}

\usepackage{textcomp,marvosym}

\usepackage{cite}

\usepackage{nameref,hyperref}

\usepackage[right]{lineno}

\usepackage{multirow}

\usepackage{microtype}
\DisableLigatures[f]{encoding = *, family = * }

\usepackage[table]{xcolor}

\usepackage{array}

\newcolumntype{+}{!{\vrule width 2pt}}

\newlength\savedwidth

\raggedright
\usepackage[aboveskip=1pt,labelfont=bf,labelsep=period,justification=raggedright,singlelinecheck=off]{caption}

\makeatletter
\renewcommand{\@biblabel}[1]{\quad#1.}
\makeatother

\usepackage{lastpage,fancyhdr,graphicx}
\usepackage{epstopdf}
\fancyheadoffset[L]{2.25in}
\fancyfootoffset[L]{2.25in}
\usepackage[normalem]{ulem}
\useunder{\uline}{\ul}{}

\begin{document}
\vspace*{0.2in}

\begin{flushleft}
{\Large
\textbf\newline{Exploring object-centric and scene-centric CNN features and their complementarity for human rights violations recognition in images} 
}
\newline
\\
Grigorios Kalliatakis\textsuperscript{1*},
Shoaib Ehsan\textsuperscript{1},
Ales Leonardis\textsuperscript{2},
Klaus McDonald-Maier\textsuperscript{1}
\\
\bigskip
\textbf{1} School of Computer Science and Electronic Engineering, University of Essex, UK
\\
\textbf{2} School of Computer Science, University of Birmingham, UK
\\
\bigskip

%
%





* gkallia@essex.ac.uk

\end{flushleft}
\section*{Abstract}
Identifying potential abuses of human rights through imagery is a novel and challenging task in the field of computer vision, that will enable to expose human rights violations over large-scale data that may otherwise be impossible. While standard databases for object and scene categorisation contain hundreds of different classes, the largest available dataset of human rights violations contains only 4 classes. Here, we introduce the `Human Rights Archive Database' (HRA), a verified-by-experts repository of 3050 human rights violations photographs, labelled with human rights semantic categories, comprising a list of the types of human rights abuses encountered at present. With the HRA dataset and a two-phase transfer learning scheme, we fine-tuned the state-of-the-art deep convolutional neural networks (CNNs) to provide human rights violations classification CNNs (HRA-CNNs). We also present extensive experiments refined to evaluate how well object-centric and scene-centric CNN features can be combined for the task of recognising human rights abuses. With this, we show that HRA database poses a challenge at a higher level for the well studied representation learning methods, and provide a benchmark in the task of human rights violations recognition in visual context. We expect this dataset can help to open up new horizons on creating systems able of recognising rich information about
human rights violations. Our dataset, codes and trained models are available online at \href{https://github.com/GKalliatakis/Human-Rights-Archive-CNNs}{https://github.com/GKalliatakis/Human-Rights-Archive-CNNs}.


\nolinenumbers


\section*{Introduction}
Human rights violations have been unfolding during the entire human history, while nowadays they increasingly appear in many different forms around the world. By `human rights violations' we refer in this paper to actions executed by state or non-state actors that breach any part of those rights that protect individuals and groups from behaviours that interfere with fundamental freedoms and human dignity \cite{United_Nations}. As mobile phones with photo and video capability are ubiquitous, individuals (human rights activists, journalists, eye witnesses and others) are recording and sharing high quality photos and videos of human rights incidents and circumstantial information. Photos and videos have become an important source of information for human rights investigations, including Commissions of Inquiry and Fact-finding Missions. Investigators often receive digital images directly from witnesses, providing high quality corroboration of their testimonies. In most instances, investigators receive images from third parties (e.g. journalists or NGOs), but their provenance and authenticity is unknown as indicated in \cite{WITNESS}. A third source of digital images is social media, e.g.\ uploaded to Facebook, again with uncertainty regarding authenticity or source \cite{McPherson}. That sheer volume of images means that to manually sift through the images to verify if any abuse is taking place and then act on it, would be tedious and time-consuming work for humans. For this reason a software tool aimed at identifying potential abuses of human rights, capable of going through images quickly to narrow down the field would greatly assist human rights investigators.

The field of computer vision has developed several databases to organize knowledge about object categories \cite{griffinHolubPerona,torralba200880,fei2007learning}, scenes \cite{zhou2014learning,xiao2010sun,zhou2016places} and materials \cite{liu2010exploring,sharan2009material,bell2015material}. However, an explicit image dataset of significant size depicting human rights violations does not currently exist. To our knowledge, the only attempt of constructing an image database in the context of human rights violations was presented in \cite{kalliatakisvisapp17}. That dataset was limited to 4 different categories of human rights violations and 100 images per category, collected by utilising manually crafted query terms. Moreover, that dataset was assembled by images available on the Internet from unverified sources and does not offer high-coverage and high-diversity of exemplars. 

In the present study, we describe in depth the construction of the \textit{Human Rights Archive (HRA)} database, and evaluate the performance of several renowned convolutional neural networks for the task of recognising human rights violations. The objective of our work is not only to compare how the features learned in object-centric CNNs and scene-centric CNNs perform, but also how well they can complement each other when used as generic features in other visual recognition tasks. We will also attempt to investigate the effects of different descriptor pooling strategies for efficient feature extraction and fusion. Finally, a visualisation of the important regions in the image for predicting target concepts, allows us to show differences in the internal representations of object-centric and scene-centric networks.

Major contributions in this work are as follows:
\begin{enumerate}
  \item A new, verified-by-experts dataset of human rights abuses, containing approximately 3k images for 8 violation categories.
  \item Assess the representation capability of deep object-centric CNNs and scene-centric CNNs for recognising human rights abuses.
  \item Attempt to enhance human rights violations recognition by combining object-centric and scene-centric CNN features over different fusion mechanisms
  \item Evaluate the effects of different feature fusion mechanisms for human rights violations recognition.
  \item A web-demo for human rights violations recognition, accessible through computer or mobile device browsers.
\end{enumerate}

The remainder of the paper is organized as follows. In Section 2 we introduce the novel Human Rights Archive database and describe its unique collection procedure based on experts-verified sources. Section 3 offers a deep insight into how transferable object-centric and scene-centric CNN features are for the task of classifying human rights violations in images, along with a web-demo for recognising human rights abuses in the wild from uploaded photos. Section 4 investigates the complementarity of object-centric and scene-centric CNN features by exploiting different fusion mechanisms. The paper concludes in Section 5. 


\section*{Human rights violations database}
\label{sec1}

At present, organizations concerned with human rights advocacy are gradually using digital images as a tool for improving the exposure of human rights and international humanitarian law violations that may otherwise be impossible. However, in order to advance the automated recognition of human rights violations a well-sampled image database is required.

\begin{figure}[!h]
\includegraphics[width=\textwidth]{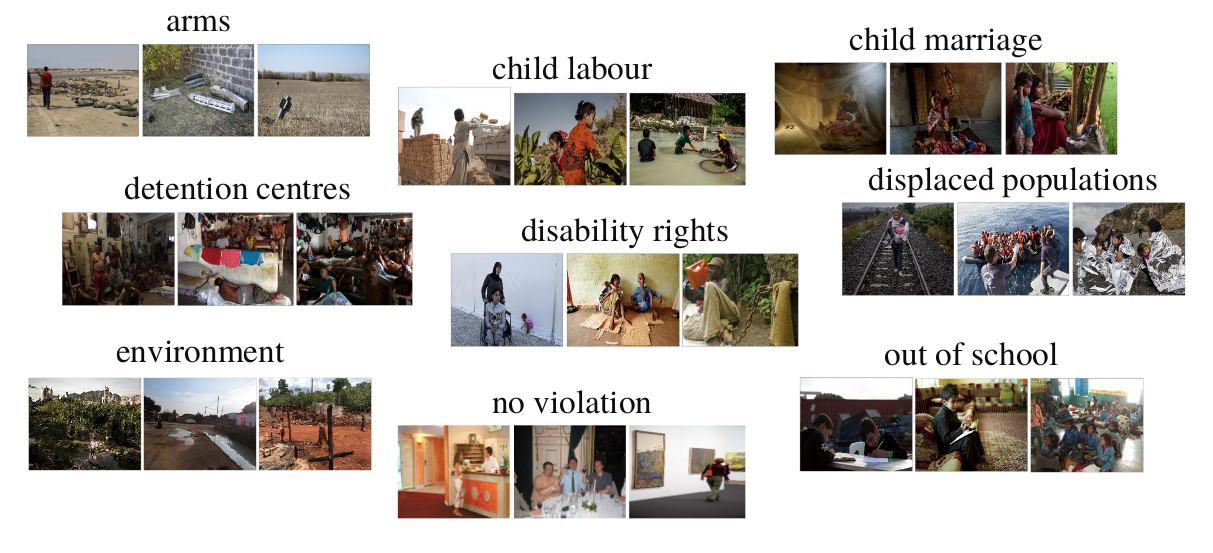}
\caption{{\bf Image samples from the categories of the Human Rights Archive (HRA) dataset.} The dataset contains eight violation categories and a supplementary  `no violation' class.} 
\label{fig1}
\end{figure}

\begin{figure}[!t]
\includegraphics[width=\textwidth]{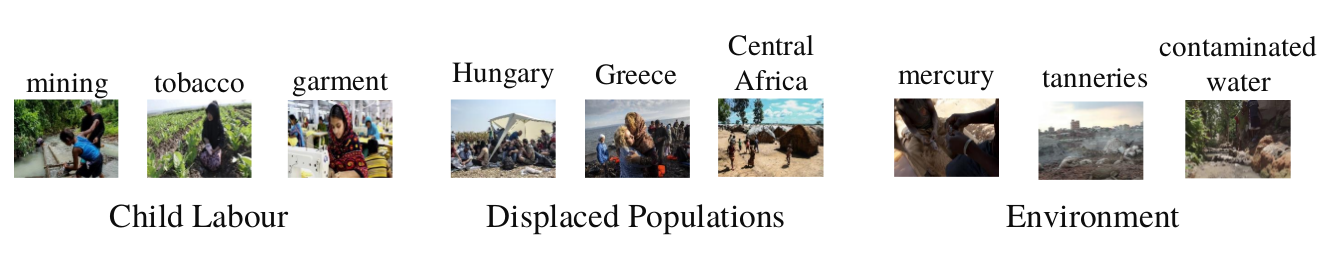}
\caption{\bf Image samples from our human rights categories grouped by different situations to illustrate the diversity of the dataset. For each situation we show 3 labelled images.} 
\label{fig2}
\end{figure}

In this section, we describe the \textit{Human Rights Archive (HRA)} database, a repository of approximately 3k well-sampled photographs of various human rights violations captured in real world situations and surroundings, labelled with 8 semantic categories, comprising the types of human rights abuses encountered in the world. Image samples are shown in Fig~\ref{fig1}. In order to increase the diversity of visual appearances in the HRA dataset (see Fig~\ref{fig2}), images from different situations or places are gathered. The dataset will be made available for future research.

\subsection*{Challenges}
\label{challenges_section}

The fundamental asset of a high-quality dataset is a broad coverage of the explicit space that needs to be learned. The intention of Human Rights Archive database is to provide a collection of human rights violations categories encountered in the world nowadays, limited to activities that can be straightforwardly utilised to answer the question of whether there is a human right being violated in an image without any other prior knowledge regarding the action. To the best of our knowledge, the largest available dataset \cite{kalliatakisvisapp17} in the context of human rights violations consists of only 4 classes and 100 images per category, with no other reference point in standardised dataset of images and annotations regarding human rights violations. The main drawback of that attempt is that the dataset was assembled by images available on the Internet from unverified sources and does not offer high-coverage and high-diversity of exemplars.

Human rights violations recognition is closely related to, but radically different from, object and scene recognition. For this reason, following a conventional image collection  procedure is not appropriate for collecting images with respect to human rights violations. The first issue encountered is that the query terms for describing different categories of human rights violations must be provided by experts in the field of human rights and not by quasi-exhaustively searching a dictionary. The next obstacle concerns online search engines such as Google, Bing or even dedicated photo-sharing websites like Flickr, which returned a huge number of irrelevant results for the given queries of human rights violations and discussed in a previous study \cite{kalliatakisHRPDA}. The final and most important matter of contention is the ground truth label verification of the images, which commonly is accomplished by crowd-sourcing the task to Amazon Mechanical Turk (AMT). However, in the case of human rights violations, human classification accuracy cannot be measured by utilising AMT for the reason that workers are not qualified for such specialised tasks.  

\subsection*{Building the Human Rights Archive database}

A key question with respect to the visual recognition problem of human rights violations from real-world images arises: \emph{how can this structured visual knowledge be gathered?} As discussed in Section \ref{challenges_section}, the crucial aspects of such a unique image database are the origin and the verification of the images. For this reason, and in order to obtain an adequate number of verified real-world images depicting human rights violations, we turn to non-governmental organizations (NGOs) and their public repositories. 

\begin{figure}[!t]
\includegraphics[width=\textwidth]{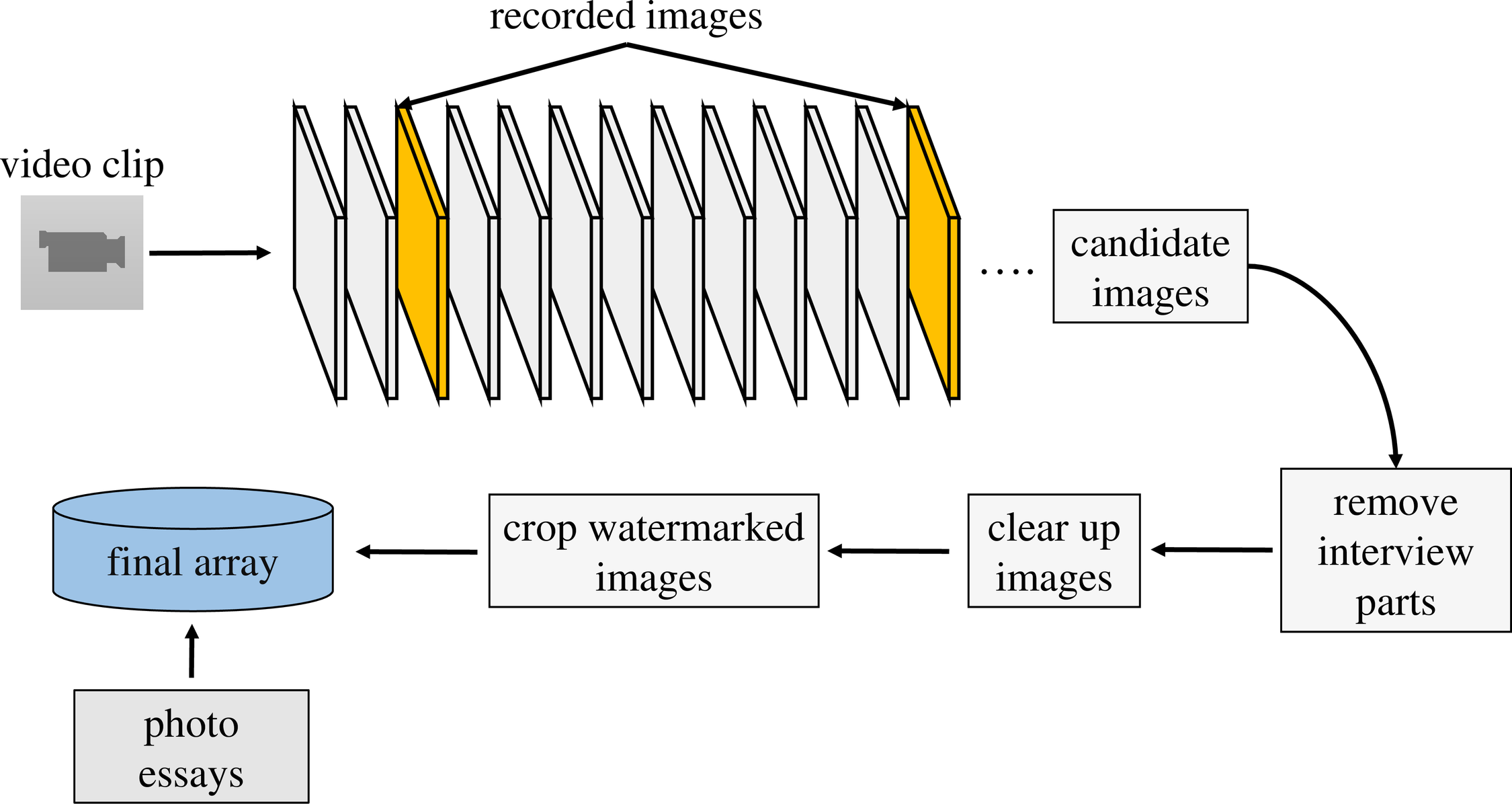}
\caption{\bf Image collection procedure from Human Rights Watch media repository.} 
\label{fig3}
\end{figure}

The first NGO considered is Human Rights Watch which offers an online media platform (\href{http://media.hrw.org/}{http://media.hrw.org/}) capable of exposing human rights and international
humanitarian law violations in the form of various media types such as videos, photo essays, satellite imagery and audio clips. Their online repository contains 9 main topics in the context of human rights violations (arms, business, children's rights, disabilities, health and human rights, international justice, LGBT, refugee rights and women rights) and 49 subcategories. In total, we download 99 available video clips from their online platform. After that, candidate images are being recorded for every video clip with a ratio of 10 (one image out of ten is recorded). This is done in order to obtain images distinctive enough on a frame to frame basis. Next, all the images that do not correspond to the definition of the human right violation category (mostly the interview parts of the clips) are manually removed. Images with low quality (very blurry or noisy, black-and-white), clearly manipulated (added text or borders, or computer-generated elements) or otherwise unusual (aerial views) are also removed. One considerable drawback in the course of that process is the presence of a watermark in most of the video files available from that platform. As a result, all the recorded images that originally contained the watermark had to be cropped in a suitable way. Only colour images of 600 x 900 pixels or larger were retrieved after the cropping stage. In addition to those images, all photo essays available for each topic and its subcategories are added, resulting in 342 more images to the final array. The entire pipeline used for collecting and filtering out the images from Human Rights Watch is depicted in Fig~\ref{fig3}.

The second NGO investigated is the United Nations which presents an online collection of images (\href{http://www.unmultimedia.org/photo/}{http://www.unmultimedia.org/photo/}) in the context of human rights. Their website is equipped with a search mechanism capable of returning relevant images for simple and complex query terms. In order to define a list of query terms, we utilise all main topics and their respective subcategories from Human Rights Watch and combine them with likely synonyms. For example, in order to acquire images depicting the employment of children in any work that deprives children of their childhood and interferes with their ability to attend regular school, `child labour', `child work' and `child employment' were provided as queries to the database. In total we download 8550 candidate images by utilising the list of query terms. We follow the same approach as Human Rights Watch in order to filter out the images. First, we manually remove all the images that do not correspond to the definition of the human right violation category. In the case of the United Nations online repository, the majority of the returned images showcased people sharing their testimony at various presentations or panel discussions. We also remove images that are black-and-white or otherwise unusual (aerial views). Finally, we add applicable high-resolution images to the database.  

\subsection*{Data analysis}

In this section, we conduct in-depth analysis in various aspects of the dataset. The final dataset contains a set of 8 human rights violations categories and 2847 images (the number of images is continuously growing as we seek additional repositories verified by other NGOs), that cover a wide range of real-world situations. 367 ready-made images are downloaded from the two online repositories representing 12.88\% of the entire dataset, while the remainder (2480) images are recorded from videos coming out of Human Rights Watch media platform. The categories are listed and defined in Table \ref{table1}. Furthermore, 203 instances which are not considered as human rights violations, such as children playing and workers mining, have been incorporated into the database in order to assess the classification performance more precisely. Our human rights-centric dataset differs from the previous Human Rights UNderstanding (HRUN) dataset \cite{kalliatakisvisapp17}. That dataset was created by collecting images available on the Internet using online search engines for different manually crafted terms, but the HRA database was created by collecting human rights violations categories from verified sources. Because it is much more difficult to find images for some human rights violations categories than for others, the distribution of images across violation categories in the database is not uniform. Examples of human rights violations categories with more images are  \textit{child labour}, \textit{displaced people}, and \textit{environment}. Examples of under-sampled categories include \textit{child marriage} and \textit{detention centres}.

\begin{table}[!t]
\begin{adjustwidth}{-2.25in}{0in} 
\centering
\caption{
{\bf Proposed human rights violations categories with definitions.}}
\begin{tabular}{|l|l|}
\hline
\textbf{1. Arms}                  & Weapons systems that put civilians at high risk of armed conflict and violence                                                                                                                                                                                                                                             \\ \hline
\textbf{2. Child Labour}          & \begin{tabular}[c]{@{}l@{}}Work that deprives children of their childhood, their potential and their dignity, \\ and that is harmful to physical and mental development\end{tabular}                                                                                                                                       \\ \hline
\textbf{3. Child Marriage}        & \begin{tabular}[c]{@{}l@{}}A formal marriage or informal union before age 18. Child marriage is widespread \\ and can lead to a lifetime of disadvantage and deprivation\end{tabular}                                                                                                                                      \\ \hline
\textbf{4. Detention Centres}     & \begin{tabular}[c]{@{}l@{}}The right to health and a healthy environment, the right to be free from discrimination \\ and arbitrary detention as critical means of achieving health\end{tabular}                                                                                                                           \\ \hline
\textbf{5. Disability Rights}     & \begin{tabular}[c]{@{}l@{}}People with disabilities experience a range of barriers to education, health care \\ and other basic services, while they are subjected to violence and discrimination\end{tabular}                                                                                                             \\ \hline
\textbf{6. Displaced Populations} & \begin{tabular}[c]{@{}l@{}}Abuses against the rights of refugees, asylum seekers, and displaced people \\ (block access to asylum, forcible return of people to places where their lives or \\ freedom would be threatened, and deprive asylum seekers of rights to fair \\ hearings of their refugee claims)\end{tabular} \\ \hline
\textbf{7. Environment}           & \begin{tabular}[c]{@{}l@{}}A lack of legal regulation and enforcement of industrial and artisanal mining, \\ large-scale dams, deforestation, domestic water and sanitation systems,\\  and heavily polluting industries can lead to host of human rights violations\end{tabular}                                          \\ \hline
\textbf{8. Out of School}         & \begin{tabular}[c]{@{}l@{}}Discrimination of marginalized groups by teachers and other students, long \\ distances to school, formal and informal school fees, and the absence of \\ inclusive education are among the main causes of children staying out of school\end{tabular}                                          \\ \hline
\end{tabular}
\label{table1}
\end{adjustwidth}
\end{table}

\subsection*{Visualising HRA}

\begin{figure}[!t]
\centering
\includegraphics[width=\textwidth,height=10cm,keepaspectratio]{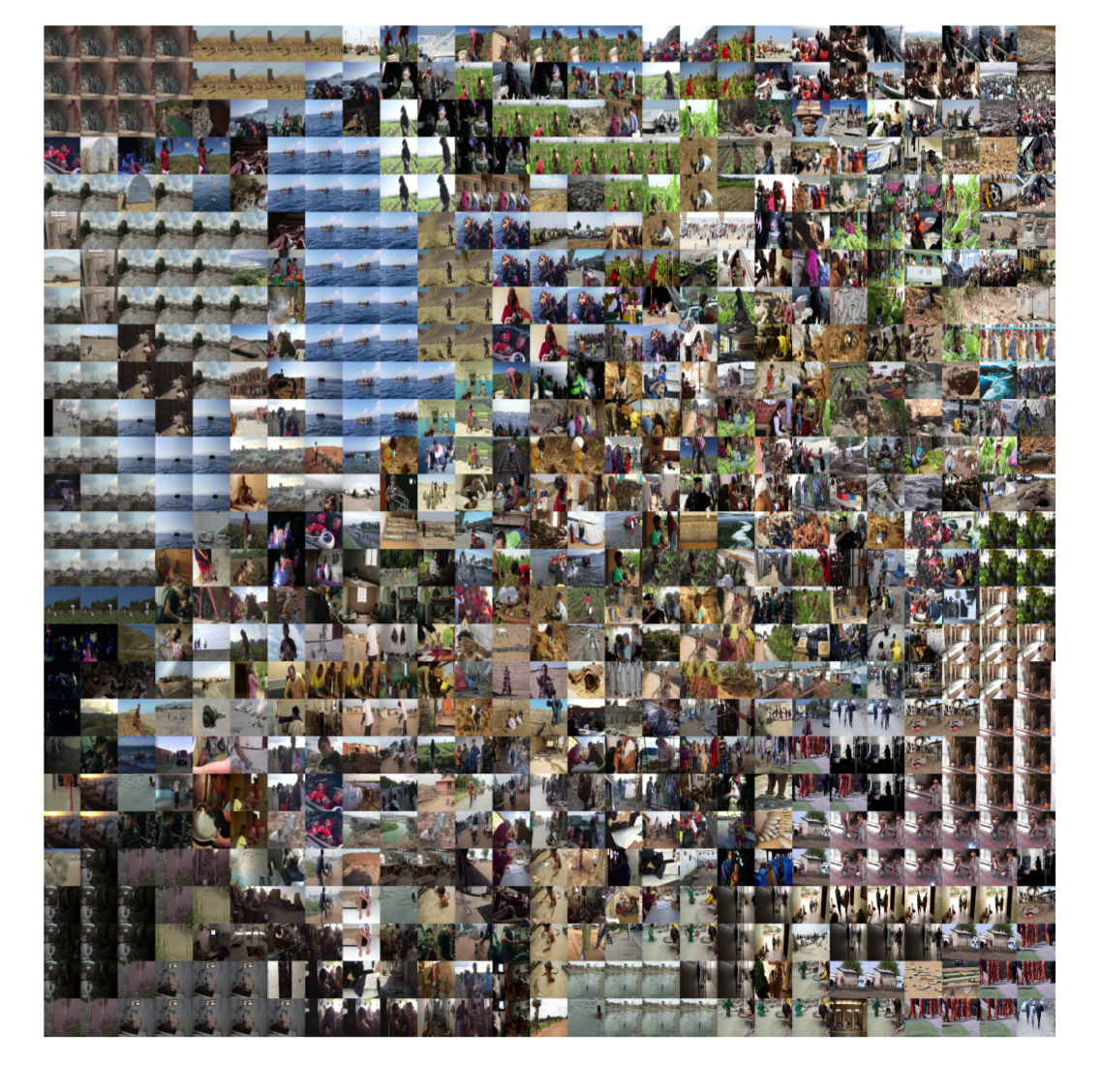}
\caption{{\bf t-SNE embedding of HRA dataset images based on their extracted features.} Images that are nearby each other are also close in the CNN representation space, which implies that the CNN `sees' them as being very similar. Notice that the similarities are more often class-based and semantic rather than pixel and colour-based.}
\label{fig4}
\end{figure}

Convolutional neural networks can be interpreted as continuously transforming the images into a representation in which the classes are separable by a linear classifier. In order to obtain an estimation about the topology of the Human Rights Archive space, we examined the internal features learned by a CNN using t-SNE (t-distributed Stochastic Neighbour Embedding) \cite{t-SNE} visualisation algorithm, by embedding images into two dimensions so that their low-dimensional representation has approximately equal distances as their high-dimensional representation. To produce that visualisation, we feed the HRA set of images through the well studied VGG-16 convolutional-layer CNN architecture \cite{VGG}, where the 4096 dimensional visual features are taken at the output of the second fully-connected layer (i.e., FC7) including the ReLU non-linearity by using caffe \cite{caffe} framework. Those features are then plugged into t-SNE in order to project the image features down to 2D. PCA preprocessing is used prior to the t-SNE routine to reduce to 10D to help optimize the t-SNE runtime. We then visualise the corresponding images in a grid as shown in Fig~\ref{fig4}, which can help us identify various clusters. Every position of the embedding is filled with its nearest neighbour. Note that since the actual embedding is roughly circular, this leads to a visualisation where the corners are a little `stretched' out and over-represented.

\section*{Convolutional neural networks for human rights violations classification}
Given the impressive classification performance of the deep convolutional neural networks, we choose three popular object-centric CNN architectures, ResNet50 \cite{ResNet} , VGG 16 convolutional-layer CNN \cite{VGG}, and VGG 19 convolutional-layer CNN \cite{VGG}, then fine-tune them on HRA to create baseline CNN models. Additionally, given the nature of the task at hand, we further fine-tuned a scene-centric CNN architecture, VGG16-Places365 \cite{zhou2017places} and compared it with the object-centric CNNs for human rights violations classification. All the HRA-CNNs presented here were trained using the Keras package \cite{chollet2015keras} on Nvidia GPU Tesla K80 (all the HRA-CNNs are available at \href{https://github.com/GKalliatakis/Human-Rights-Archive-CNNs}{https://github.com/GKalliatakis/Human-Rights-Archive-CNNs}. 

\begin{figure}[!t]
\includegraphics[width=\textwidth,keepaspectratio]{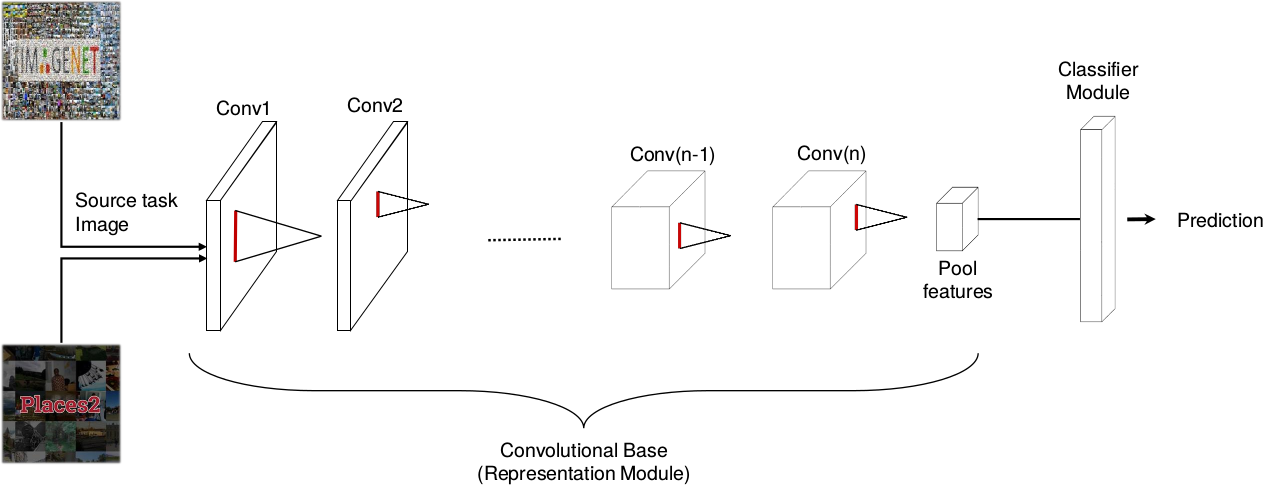}
\caption{{\bf General structure of CNN architecture for end-to-end image classification.}}
\label{fig5}
\end{figure}

The selected CNN architectures contain 138 million parameters for VGG16, 143 million parameters for VGG19 and 26 million parameters for ResNet50. VGG16-Places365 and VGG16 have exactly the same network architecture, but they are trained on scene-centric data and object-centric data respectively. Directly learning so many parameters from only a few thousand training images is problematic. A general structure of CNN architecture is depicted in Fig~\ref{fig5}. 

\subsection*{Transferring CNN weights}
\label{transfer_learning}
A conventional approach to enable training of very deep networks on relative small datasets is to use a model pre-trained on a very large dataset, and then use the CNN as as an initialization for the task of interest. This method, referred to as `transfer learning' \cite{transfer_learning, DeCAF, Zeiler} injects knowledge from other tasks by deploying weights and parameters from a pre-trained network to the new one \cite{kalliatakis_materials} and has become a commonly used method to learn task-specific features. The key idea is that the internal layers of the CNN can act as a generic extractor of high-level image representations, which can be pre-trained on one large dataset, the source task, and then re-used on other target tasks \cite{oquab2014learning}. Considering the size of our dataset the chosen method to apply a deep CNN is to reduce the number of free parameters. In order to achieve this, the first filter stages can be trained in advance on different tasks of
object or scene recognition and held fixed during training on human rights violations recognition. By freezing (preventing the weights from getting updated during training) the earlier layers, overfitting can be avoided. We initialize the feature extraction modules using pre-trained models from two different large scale datasets, ImageNet \cite{Krizhevsky} and Places\cite{zhou2017places}. ImageNet is an object-centric dataset which contains images of generic objects including person and
therefore is a good option for understanding the contents
of the image region comprising the target person. On the
contrary, Places is a scene-centric dataset specifically created for high
level visual understanding tasks such as recognizing scene
categories. Hence, pretraining the image feature extraction model using this dataset ensures providing global (high level) contextual support. For the target task (human rights violation recognition), we design a network that will output scores for the eight target categories of the HRA dataset or {\fontfamily{pcr}\selectfont no violation} if none of the categories are present in the image.

\begin{figure}[!t]
\includegraphics[width=\textwidth,keepaspectratio]{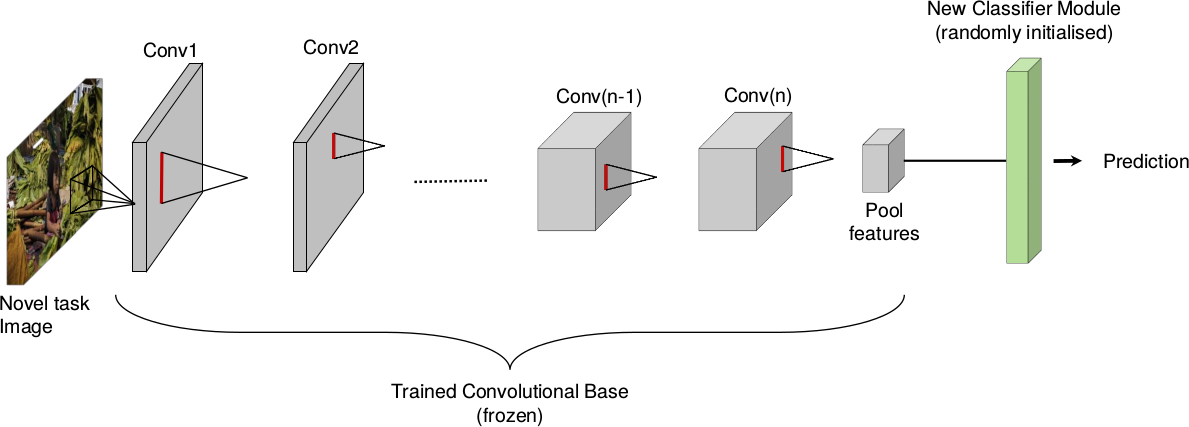}
\caption{{\bf Network architecture used for high-level feature extraction with HRA.} Pre-trained parameters of the internal layers
of the networks are transferred to the target task. To compensate for the different nature of the source and target data we add a randomly initialised adaptation layer (fully connected layer) and train them on the labelled data of the target task.}
\label{fig6}
\end{figure}

Transfer is achieved in two phases. First, we start by using the representations learned by a previous network in order to extract interesting features from new samples. `Feature extraction' consists of taking the convolutional base of a pre-trained network, running the new data of HRA through it and training a new, randomly initialised classifier on top of the semantic image output vector $\textbf{Y}_{out}$, as illustrated in Fig~\ref{fig6}. Note that $\textbf{Y}_{out}$ is obtained as a complex non-linear function of potentially all input pixels and captures the high-level configurations of objects or scenes. We intentionally utilise only the convolutional base and not the densely-connected classifier of the original network, merely because the representations learned by the convolutional base are likely to be more generic. More importantly, representations found in densely-connected layers no longer contain any spatial information which is relevant for the task at hand. Note that in our experiments, the pooling layer just before the new classifier can be either a global average/max pooling operation for spatial data or simply a flattening layer. The $FC_{HRA}$  layer compute $\textbf{Y}_{HRA} = \sigma(\textbf{W}_{HRA}{\textbf{Y}}_{out} + \textbf{B}_{HRA} )$, where \textbf{W}, \textbf{B} are the trainable parameters. In all our experiments, the last convolutional layer of the pre-trained base have sizes (7, 7, 512). 

\begin{figure}[!t]
\includegraphics[width=\textwidth,keepaspectratio]{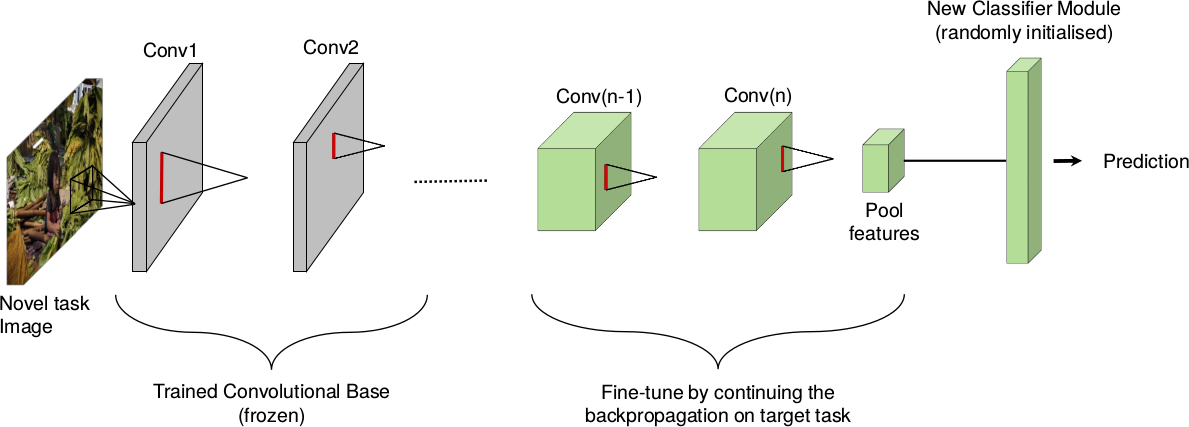}
\caption{{\bf Network architecture used for fine-tuning with HRA.} It marginally alters the more abstract representations of the model being utilised, in order to make them more relevant for the problem at hand.}
\label{fig7}
\end{figure}

The second phase required for transfer learning, complementary to feature extraction, is \textit{fine-tuning}. Fine-tuning consists of unfreezing few of the top layers of a previously frozen convolutional base for feature extraction, and jointly training both the newly added fully-connected classifier and these top layers as illustrated in Fig~\ref{fig7}. It is only beneficial to fine-tune the top layers of the convolutional base once the classifier on top has already been trained (see Fig~\ref{fig6}). This is because the large gradient updates triggered by the randomly initialized weights would wreck the learned weights in the convolutional base. We choose to only fine-tune the last two convolutional layers rather than the entire network in order to prevent overfitting, since the entire network would have a very large entropic capacity and thus a strong tendency to overfit. The features learned by low-level convolutional layers are more general,
less abstract than those found higher-up, so it is sensible to keep the first
few layers fixed (more general features) and only fine-tune the last two (more specialized features).

For all of our experiments, we use the HRA dataset exclusively for the training process, while we obtain other representative images for each category from the Internet in order to compose the test set, producing a total of 270 reasonable images. Thus we eliminate the presence of bias in our experiments while our models are tested in the wild with real-world images. Table \ref{table2} summarises the statistics of the HRA dataset. For the purposes of our experiments, the data is divided into two main subsets: training/validation data (trainval), and test data (test). To compensate for the imbalanced classes in HRA, we utilise cost-sensitive training to weight the loss function during training by an amount proportional to how under-represented each class is. This is useful to tell the model to `pay more attention' to samples from an under-represented class. The maximum number of epochs was set to 40  iterations for each epoch and a learning rate of 0.0001, using the stochastic gradient descent (SGD) optimizer for cross-entropy minimization. The parameters were chosen empirically by analysing the training loss.  

\begin{table}[!b]
\centering
\caption{{\bf Statistics of the HRA dataset.} The data is divided
into two main subsets: training/validation data (trainval),
and test data (test), with the trainval data further divided
into suggested training(train) and validation (val) sets.}
\begin{tabular}{ccccc}
\hline
\multirow{2}{*}{\textbf{}} & train  & val    & trainval & test   \\ \cline{2-5} 
                           & images & images & images   & images \\ \hline
arms                       & 149    & 37     & 186      & 30     \\
child labour               & 756    & 189    & 945      & 30     \\
child marriage             & 69     & 18     & 87       & 30     \\
detention centres          & 149    & 37     & 186      & 30     \\
disability rights          & 218    & 55     & 273      & 30     \\
displaced populations      & 487    & 122    & 609      & 30     \\
environment                & 326    & 82     & 408      & 30     \\
no violation               & 162    & 41     & 203      & 30     \\
out of school              & 123    & 30     & 153      & 30     \\[0.15cm]
Total                      & 2439   & 611    & 3050     & 270    \\ \hline
\end{tabular}
\label{table2}
\end{table}

\subsection*{Results on Human Rights Archive (HRA)}

After fine-tuning the various CNNs, we used the final output layer of each network to classify the test set images of HRA. The classification results for \textit{top-1 accuracy} and \textit{coverage} are listed in Table \ref{table3}. The  is the fraction of correctly classified samples from the test set, where the top predicted label exactly match the ground-truth label. In the case of human rights violations recognition we are also interested in how confident a model should be about a decision, especially because a human operator is supposed to take over. If the system thinks that it is less likely than a human being to obtain correct interpretation, then the best route of action is to allow a human to interpret the photo instead. A more natural performance metric to use in this situation is \textit{coverage}, the fraction of examples for which the system is able to produce a response. For all the experiments in this paper we employ a threshold of 0.85 over the prediction confidence in order to report the coverage performance metric.

\begin{figure}[!t]
\includegraphics[width=\textwidth,keepaspectratio]{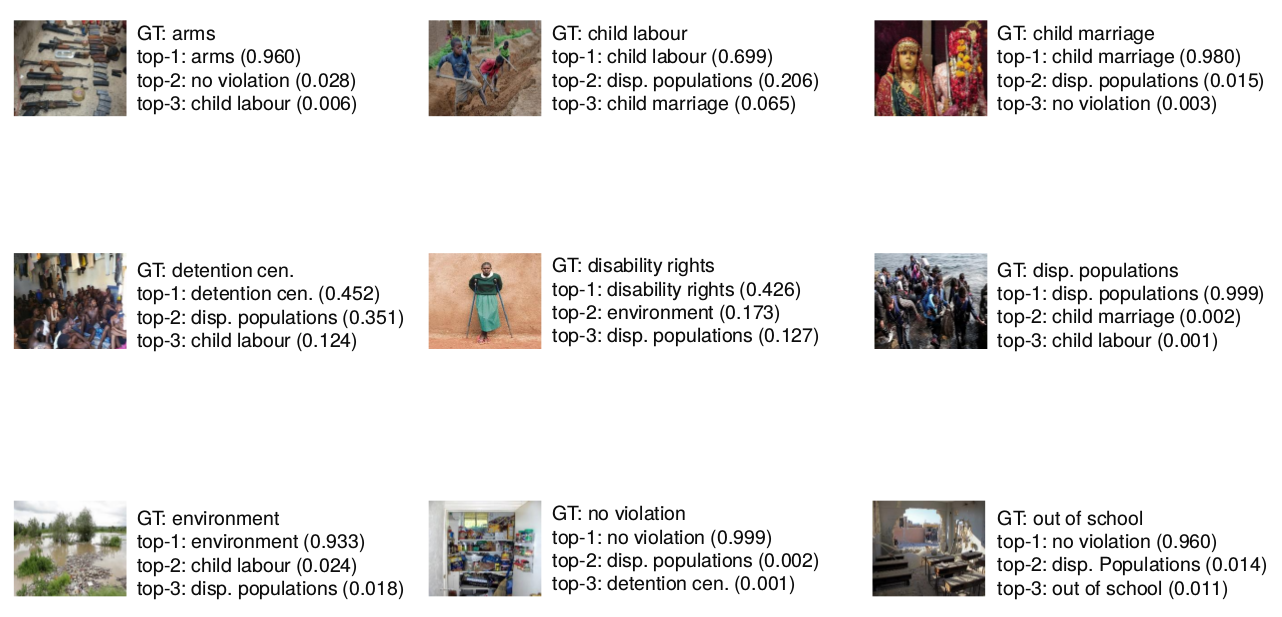}
\caption{{\bf The predictions given by the best performing HRA-VGG19 for the images from the test set.} The ground-truth label (GT) and the top 3 predictions are shown. The number beside each label indicates the prediction confidence.}
\label{fig8}
\end{figure}

Fig~\ref{fig8} shows the responses to examples predicted by the best performing HRA-CNN, VGG19. Broadly, we can identify one type of misclassification given the current label attribution of HRA: images depicting the evidence
which are responsible for a particular situation and not the actual action, such as schools being targeted by armed attacks. Future development of the HRA database, will explore to assign multi-ground truth
labels or free-form sentences to images to better capture the richness of visual descriptions of human rights violations.

\begin{table}[!b]
\centering
\caption{{\bf Classification accuracy/precision and coverage on the test set of HRA for the deep features of various object-centric CNNs (the first three models) and a scene-centric CNN (the last model).} Bold font highlights the dominant performance across the same metric.}
\begin{tabular}{ccccc}
\hline
                & Pool                     & Top-1 acc.       & Coverage      & Trainable Params. \\ \hline
VGG16           & \multirow{4}{*}{avg}     & 34.44\%          & 45\%          & 4,853,257         \\
VGG19           &                          & \textbf{35.18\%} & 42\%          & 4,853,257         \\
ResNet50        &                          & 25.55\%          & 55\%          & 4,992,521         \\
VGG16-places365 &                          & 30.00\%          & 32\%          & 4,853,257         \\ \hline
VGG16           & \multirow{4}{*}{flatten} & 31.85\%          & 55\%          & 8,784,905         \\
VGG19           &                          & 31.11\%          & 50\%          & 8,784,905         \\
ResNet50        &                          & 30.00\%          & 44\%          & 4,992,521         \\
VGG16-places365 &                          & 28.51\%          & 52\%          & 8,784,905         \\ \hline
VGG16           & \multirow{4}{*}{max}     & 28.14\%          & \textbf{64\%} & 4,853,257         \\
VGG19           &                          & 29.62\%          & 61\%          & 4,853,257         \\
ResNet50        &                          & 25.55\%          & 61\%          & 4,992,521         \\
VGG16-places365 &                          & 26.66\%          & 51\%          & 4,853,257         \\ \hline
\end{tabular}
\label{table3}
\end{table}

We can see that both VGG architectures surpass the scene-centric architecture of VGG16-Places365 by a significant margin of at least 4.44\% for top-1 accuracy and 10\% for coverage for their best performing pooling operation, even though the number of trainable parameters remains exactly the same. On the other hand, VGG16-Places365 outperform the object-centric ResNet50 for two of the pooling schemes. We have also tried to change the number of layers which were fine-tuned in our training set-up. Increasing the number of layers to three results in about 7\% drop in classification performance. It is evident from Table \ref{table3} that each object-centric and scene-centric CNN has different strengths and weaknesses. Therefore, we expect that using an ensemble of different models would further boost the accuracy for the task of recognising human rights violations.

\subsection*{Model interpretation}
\label{model_interpetation_section}
\begin{figure}[!t]
\includegraphics[width=\textwidth,keepaspectratio]{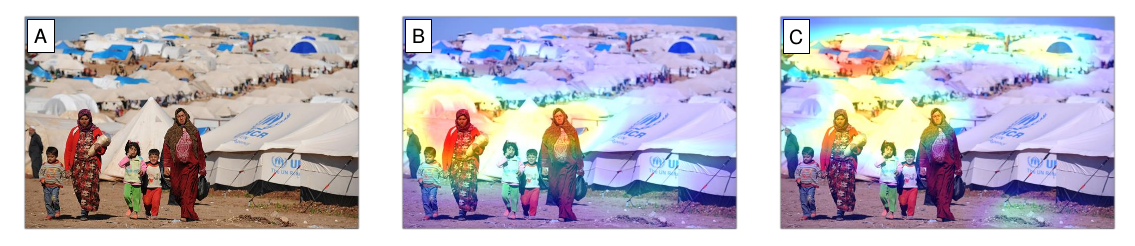}
\caption{{\bf Given an input image A, we visualise the class-discriminative regions of different CNNs using Grad-CAM  \cite{selvaraju2016grad} for the
output class ‘displaced populations’.} The object-centric model \textbf{B} focuses on the head of the people, while the scene-centric model \textbf{C} focuses on the shelters in the background.}
\label{fig9}
\end{figure}

In order to interpret which parts of a given image led a CNN to its final prediction, we produce heatmaps of  `class activation'. Class Activation Mapping (CAM) \cite{zhou2016learning} and its generalisation Gradient-weighted Class Activation Mapping (Grad-CAM) \cite{selvaraju2016grad} visualise the linear activations of a late layer's activations with respect to the class considered. To generate Grad-CAM visual explanations, we followed the approach of \cite{selvaraju2016grad}. An image is fed into the fine-tuned network and the output feature maps of the last convolutional layer are extracted. Convolutional features are capable of retaining spatial information compared to fully-connected layers where that information is lost. The gradient of the score associated with a specific output class is computed, with respect to the extracted feature maps of the last convolutional layer. Then, the gradients are global-average-pooled to obtain the importance weights. Finally, the Grad-CAM is obtained by performing a weighted combination of forward activation maps followed by a ReLU. Fig~\ref{fig9} shows an example of Grad-CAMs for the output class of `displaced populations' .

\subsection*{Web-based software for human rights violations recognition}
\label{demo_section}
\begin{figure}[!t]
\includegraphics[width=\textwidth,keepaspectratio]{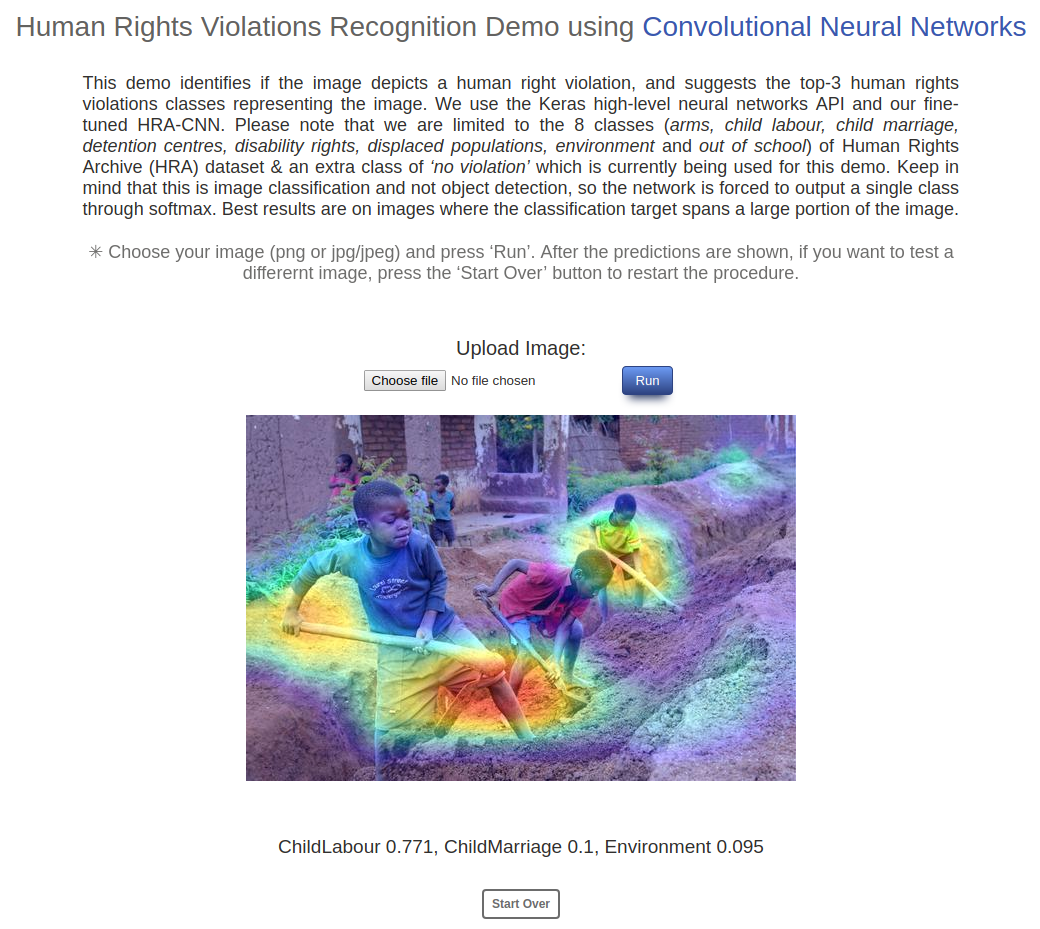}
\caption{{\bf A screenshot of the human rights violations recognition demo based on the fine-tuned HRA-CNN.} The web-demo predicts the type of human right that is being violated for uploaded photos.}
\label{fig10}
\end{figure}

Based on our trained HRA-CNNs, we created a web-demo for human rights violations recognition \href{http://83.212.117.19:5000/}{http://83.212.117.19:5000/}, accessible through computer or mobile device browsers. It is possible to upload photos to the web-based software to identify if images depict a human right violation, while the system suggests the 3 most likely semantic categories from the HRA dataset. A screenshot of the prediction result on a web browser is shown in Fig~\ref{fig10}. More precisely, the Keras \cite{chollet2015keras} python deep learning framework over TensorFlow \cite{abadi2016tensorflow} was used to train the back-end prediction model in the demo. With this system, those combating abuse will be able to go through images very quickly to narrow down the field and identify pictures which need to be looked at in more detail. Furthermore, with the extensive use of this software, we will collect an expanded range of images depicting human rights abuses, in order to enhance the accuracy of our CNN models with larger data sets. Future directions for this work will include the capacity to receive feedback from people regarding the result. The source code for the web-based software is available at \href{https://github.com/GKalliatakis/Human-Rights-Archive-CNNs}{https://github.com/GKalliatakis/Human-Rights-Archive-CNNs} to assist future research.

\section*{Combining scene-centric and object-centric CNNs}
Information fusion can be a crucial component in image classification schemes where increasing the overall accuracy of the system is regarded as one of its most integral aspects. Merging different information is not only meaningful because of the accuracy improvement it can offer in a system, but also for allowing the system to be more robust against changing dynamics. Feature fusion accommodates features coming from different sources into a single representation and has been an active area of research for computer vision researchers during the last few years \cite{mangai2010survey}. Inspired by the observation that the task of recognising human rights violations is fundamentally different from the familiar tasks of object/scene recognition, we adopt two fusion concepts with the purpose of investigating how well internal representations learned by object-centric and scene-centric networks complement each other and whether an ensemble of classifiers can improve the performance on the human rights violations recognition problem.

\begin{figure}[!t]
\includegraphics[width=\textwidth,keepaspectratio]{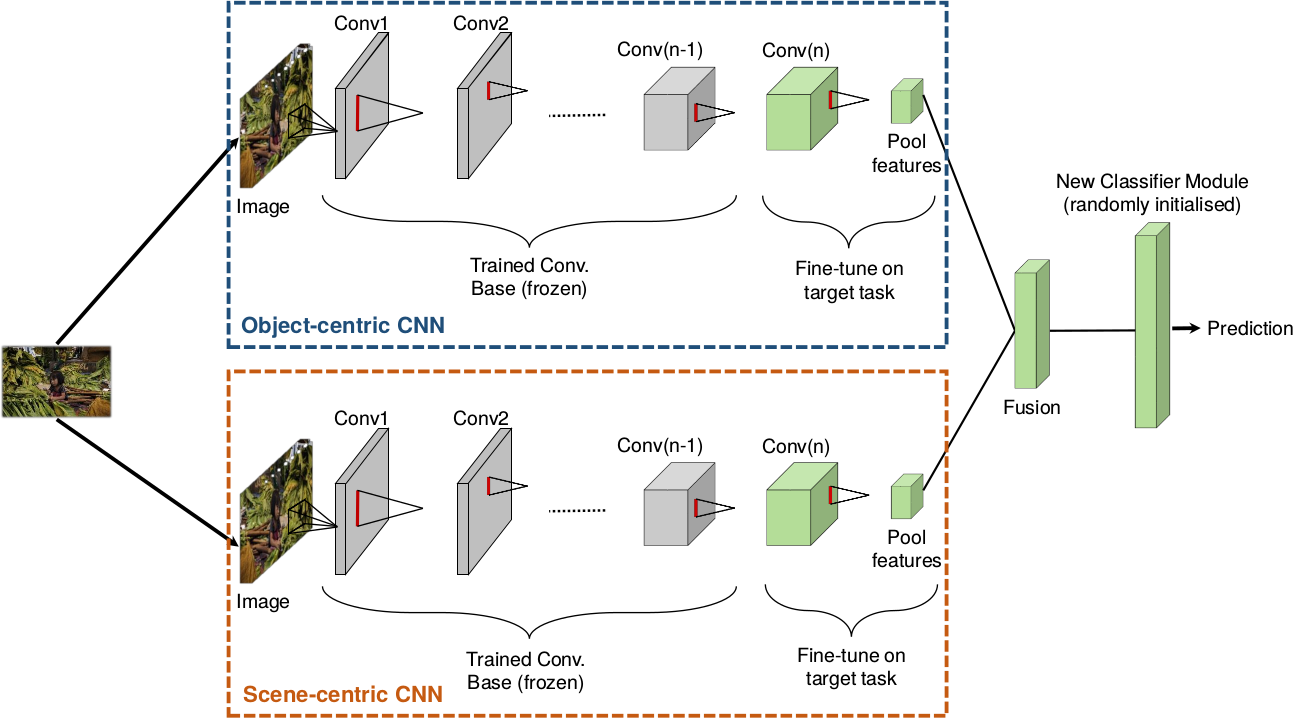}
\caption{{\bf Early fusion.} }
\label{fig11}
\end{figure}

As indicated by \cite{snoek2005early,ergun2016early}, feature fusion approaches can be grouped into two main categories: \textit{early} and \textit{late} fusion. We employ early and late fusion schemes in different ways along with the CNN architectures. The processing pipelines of our early and late fusion schemes are depicted in Fig~\ref{fig11} and Fig~\ref{fig12} respectively.

\subsection*{Proposed fusion schemes}
\subsubsection*{Early fusion}
Suppose we are given two CNNs, an object-centric network and a scene-centric network. Let feature set  $ F = \{f_1,f_2\}$ be extracted features of the last convolutional layer from each network, where each feature is a high-dimensional feature vector represented with $f_i\in \mathbb{R}^{d_i } $. Every distinct feature may have different cardinality according to the particular CNN architecture, such that $d_i = \{d_1,d_2\} $. Then feature fusion function $\phi$ can be defined as the mapping operator on \textit{F} such that $ \phi(F) \mapsto R^d  $.

The first strategy exploited in the early fusion scheme is the \textit{concatenation} method, where discrete feature vectors of different sources are concatenated into one super-vector $f_f = \{f_1,f_2\} $ which will represent the final image feature. The final vector size is the summation of all feature dimensions $ d = \sum_{i=1}^{n} d_i $ .

The second fusion strategy employed is \textit{averaging}, also known as \textit{sum pooling} in the context of neural networks. In this strategy, the feature set \textit{F} is averaged in order to form the final image descriptor $f_f = \frac{1}{n} \sum_{i=1}^{n} f_i $. All features in \textit{F} should either have the same cardinality or the feature dimensions must be normalised prior to fusion operation.

The last fusion strategy utilised is \textit{maximum pooling}. It involves the same preprocessing in terms of the final feature cardinality, however it varies in the way features are merged. Maximum pooling selects the highest value from the corresponding features instead of taking the average of all features elements' in sum pooling. If the final feature representation is $ f_f \mapsto R^d  $, then max pooling selects each member of $ f_f$ as $ f_f^i = \max_{i=1}^{d}(f_1^i, f_2^i)$.

\subsubsection*{Late fusion}
\begin{figure}[!t]
\includegraphics[width=\textwidth,keepaspectratio]{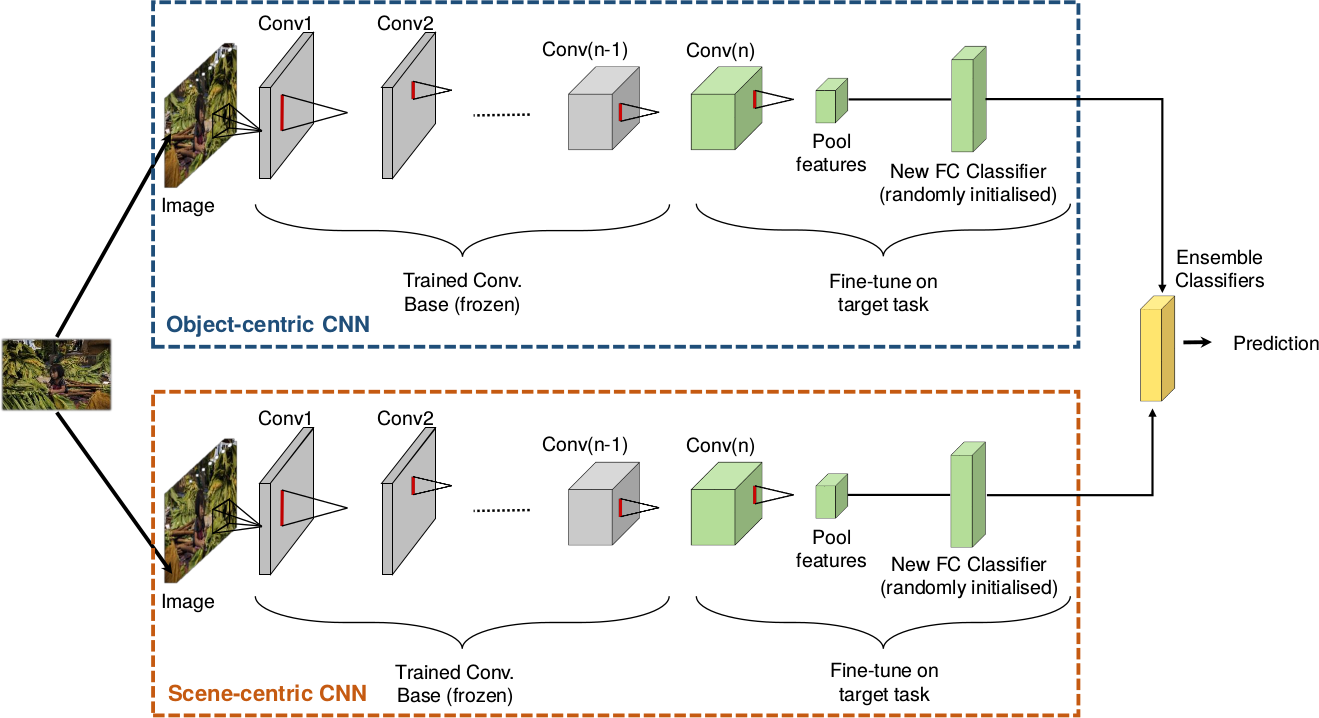}
\caption{{\bf Late fusion.} }
\label{fig12}
\end{figure}

Contrary to early fusion, the late fusion scheme consists of pooling together the predictions of a set of different end-to-end models (in our case object-centric and scene-centric CNNs), to produce more accurate predictions. This kind of assemblage relies on the assumption that independently trained object-centric and scene-centric models are focusing at slightly different aspects of the data to make their predictions as illustrated in Fig~\ref{fig9}. The easiest way to pool the predictions of a set of classifiers is to average their predictions at inference time as illustrated by Fig~\ref{fig12}.

\subsection*{Differences and complementarities}

After evaluating deep features of various object-centric and scene-centric CNNs on the test set of HRA, we turn our attention to the problem of combining those features for the same task. First, we start by transferring CNN weights as described previously, this time combining the output of the last convolutional layer of an object-centric CNN with the output of a scene-centric CNN before randomly initialising a new fully-connected classifier as shown in Fig~\ref{fig11}. Note that in this approach only the last convolutional layer of each network is fine-tuned in order to keep equal number of trainable parameters with the previous set-up. We compare results of three fusion and pooling operations and their combinations as illustrated in Table \ref{table4}.

\begin{table}[!ht]
\begin{adjustwidth}{-2.25in}{0in} 
\centering
\caption{{\bf Classification accuracy/precision and coverage on the test set of HRA using early fusion of deep features.} There are two columns for each performance metric for reporting the benefit (denoted with the (+) symbol) or the loss (denoted with the (-) symbol) of the merged model compared with the corresponding individual one. Bold font highlights the dominant performance across the same metric.}
\begin{tabular}{cccccccccc}
\hline
                                                                                        & \multirow{2}{*}{Pool}    & \multirow{2}{*}{Fusion} & \multicolumn{3}{c}{Accuracy}                                                                                                    & \multicolumn{3}{c}{Coverage}                                                                                                 & \multirow{2}{*}{Trainable Params.} \\ \cline{4-9}
                                                                                        &                          &                         & Top-1            & \begin{tabular}[c]{@{}c@{}}vs.\\ object\end{tabular} & \begin{tabular}[c]{@{}c@{}}vs.\\ scene\end{tabular} & Fraction      & \begin{tabular}[c]{@{}c@{}}vs.\\ object\end{tabular} & \begin{tabular}[c]{@{}c@{}}vs.\\ scene\end{tabular} &                                    \\ \hline
\multirow{9}{*}{\begin{tabular}[c]{@{}c@{}}VGG16\\ +\\ VGG16-places365\end{tabular}}    & \multirow{3}{*}{avg}     & average                 & 31.48\%          & −2.96\%                                               & +1.48\%                                              & 14\%          & −31\%                                                 & −18\%                                                & 4,853,257                          \\
                                                                                        &                          & concatenate             & 30.37\%          & −4.07\%                                               & +0.37\%                                              & 25\%          & −20\%                                                 & −7\%                                                 & 4,984,329                          \\
                                                                                        &                          & maximum                 & 30.74\%          & −3.70\%                                               & +0.74\%                                              & 19\%          & −26\%                                                 & −13\%                                                & 4,853,257                          \\ \cline{2-10} 
                                                                                        & \multirow{3}{*}{flatten} & average                 & 27.40\%          & −4.45\%                                               & −1.11                                                & 57\%          & +2\%                                                  & +5\%                                                 & 11,144,713                         \\
                                                                                        &                          & concatenate             & 27.77\%          & −4.08\%                                               & −0.74                                                & 58\%          & +3\%                                                  & +6\%                                                 & 17,567,241                         \\
                                                                                        &                          & maximum                 & 29.25\%          & −2.6\%                                                & +0.74\%                                              & 54\%          & -1\%                                                  & +2\%                                                 & 11,144,713                         \\ \cline{2-10} 
                                                                                        & \multirow{3}{*}{max}     & average                 & 25.18\%          & −2.96\%                                               & −1.48                                                & 45\%          & −19\%                                                 & −6\%                                                 & 4,853,257                          \\
                                                                                        &                          & concatenate             & 27.40\%          & −0.74\%                                               & +0.74\%                                              & 49\%          & −15\%                                                 & −2\%                                                 & 4,984,329                          \\
                                                                                        &                          & maximum                 & 24.44\%          & −3.7\%                                                & −2.22                                                & 56\%          & −8\%                                                  & +5\%                                                 & 4,853,257                          \\ \hline
\multirow{9}{*}{\begin{tabular}[c]{@{}c@{}}VGG19\\ +\\ VGG16-places365\end{tabular}}    & \multirow{3}{*}{avg}     & average                 & 27.03\%          & −8.15                                                 & −2.97                                                & 14\%          & −28\%                                                 & −18\%                                                & 4,853,257                          \\
                                                                                        &                          & concatenate             & 27.77\%          & −7.41                                                 & −2.23                                                & 25\%          & −17\%                                                 & −7\%                                                 & 4,984,329                          \\
                                                                                        &                          & maximum                 & 28.88\%          & −6.3                                                  & −1.12                                                & 28\%          & −14\%                                                 & −4\%                                                 & 4,853,257                          \\ \cline{2-10} 
                                                                                        & \multirow{3}{*}{flatten} & average                 & 25.55\%          & −5.56                                                 & −2.96                                                & \textbf{64\%} & \textbf{+14\%}                                        & \textbf{+12\%}                                       & 11,144,713                         \\
                                                                                        &                          & concatenate             & 28.88\%          & −2.22                                                 & +0.37\%                                              & 50\%          & 0\%                                                   & −2\%                                                 & 17,567,241                         \\
                                                                                        &                          & maximum                 & 28.14\%          & −2.97                                                 & −0.37                                                & 51\%          & +1\%                                                  & −1\%                                                 & 11,144,713                         \\ \cline{2-10} 
                                                                                        & \multirow{3}{*}{max}     & average                 & 27.03\%          & −2.59                                                 & +0.37\%                                              & 37\%          & −24\%                                                 & −14\%                                                & 4,853,257                          \\
                                                                                        &                          & concatenate             & 26.29\%          & −3.33                                                 & −0.37                                                & 47\%          & −14\%                                                 & −4\%                                                 & 4,984,329                          \\
                                                                                        &                          & maximum                 & 26.29\%          & −3.33                                                 & −0.37                                                & 48\%          & −13\%                                                 & −3\%                                                 & 4,853,257                          \\ \hline
\multirow{9}{*}{\begin{tabular}[c]{@{}c@{}}ResNet50\\ +\\ VGG16-places365\end{tabular}} & \multirow{3}{*}{avg}     & average                 & 27.03\%          & +1.48\%                                               & −2.97                                                & 5\%           & −50\%                                                 & −27\%                                                & 2,494,473                          \\
                                                                                        &                          & concatenate             & 28.51\%          & +2.96\%                                               & −1.49                                                & 14\%          & −41\%                                                 & −18\%                                                & 2,625,545                          \\
                                                                                        &                          & maximum                 & \textbf{31.85\%} & +6.3\%                                                & +1.85\%                                              & 16\%          & −39\%                                                 & −16\%                                                & 2,494,473                          \\ \cline{2-10} 
                                                                                        & \multirow{3}{*}{flatten} & average                 & 27.77\%          & −2.23                                                 & −0.74                                                & 41\%          & −3\%                                                  & −11\%                                                & 8,785,929                          \\
                                                                                        &                          & concatenate             & 27.03\%          & −2.97                                                 & −1.48                                                & 50\%          & +6\%                                                  & −2\%                                                 & 15,208,457                         \\
                                                                                        &                          & maximum                 & 24.81\%          & −5.19                                                 & −3.7                                                 & 50\%          & +6\%                                                  & −2\%                                                 & 8,785,929                          \\ \cline{2-10} 
                                                                                        & \multirow{3}{*}{max}     & average                 & 25.92\%          & +0.37\%                                               & −0.74                                                & 34\%          & −27\%                                                 & −17\%                                                & 2,494,473                          \\
                                                                                        &                          & concatenate             & 25.55\%          & 0                                                     & −1.11                                                & 41\%          & −20\%                                                 & −10\%                                                & 2,625,545                          \\
                                                                                        &                          & maximum                 & 31.11\%          & \textbf{+5.56\%}                                      & \textbf{+4.45\%}                                     & 30\%          & −31\%                                                 & −21\%                                                & 2,494,473                          \\ \hline
\end{tabular}
\label{table4}
\end{adjustwidth}
\end{table}

\begin{figure}[!b]
\includegraphics[width=\textwidth,keepaspectratio]{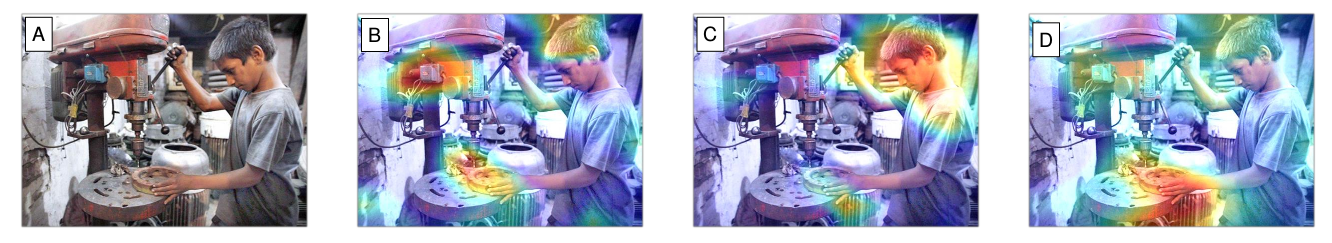}
\caption{{\bf Informative region for predicting the category ‘child labour’ for different CNNs.} Given an input image \textbf{A}, we visualise the class-discriminative regions using Grad-CAM  \cite{selvaraju2016grad} for the output class. The object-centric model \textbf{B} focuses on the tool used by the young boy, the scene-centric model \textbf{C} focuses mostly on the head of the young boy, while the early fusion of the two CNNs \textbf{D} focuses more on what the young boy is holding.}
\label{fig13}
\end{figure}

Remarkably, results indicate that early fusion of object-centric and scene-centric features constantly trail their individual counterparts in most of the evaluations for both performance metrics. More precisely, the global best coverage of 64\% (VGG19 with average pooling) can be matched by \textit{VGG19+VGG16-places365} when features were flattened and average fusion was utilised. However, the global best accuracy of 35.18\% surpass all early fusion schemes by a significant margin of at least 3.33\%. An interesting observation is that object-centric features complement profitably the accuracy of their scene-centric counterparts in numerous combinations. Through the visualisation of the class-discriminative regions in Fig~\ref{fig13}, we can have a better understanding of what has been learned inside the CNNs for the early fusion scheme.

Regarding late fusion, although the ensemble classifier of object-centric and scene-centric models fall behind their individual counterparts in most of the evaluations, late fusion consistently outperforms the early fusion scheme in accuracy. This is mostly due to lower cardinality of trainable parameters.

\begin{table}[!t]
\begin{adjustwidth}{-2.25in}{0in} 
\centering
\caption{{\bf Classification accuracy/precision and coverage on the test set of HRA using late fusion (ensemble the classifiers).} There are two columns for each performance metric for reporting the benefit (denoted with the (+) symbol) or the loss (denoted with the (-) symbol) of the merged model compared with the corresponding individual one. Bold font highlights the dominant performance across the same metric.}
\begin{tabular}{cccccccc}
\hline
                         & \multirow{2}{*}{Pool}    & \multicolumn{3}{c}{Accuracy}                                                                                                  & \multicolumn{3}{c}{Coverage}                                                                                               \\ \cline{3-8} 
                         &                          & Top-1            & \begin{tabular}[c]{@{}c@{}}vs.\\ object\end{tabular} & \begin{tabular}[c]{@{}c@{}}vs.\\ scene\end{tabular} & Fraction      & \begin{tabular}[c]{@{}c@{}}vs.\\ object\end{tabular} & \begin{tabular}[c]{@{}c@{}}vs.\\ scene\end{tabular} \\ \hline
VGG16+VGG16-places365    & \multirow{3}{*}{avg}     & \textbf{32.92\%} & −1.52                                                & \textbf{+2.92\%}                                    & 31\%          & −14\%                                                & \textbf{−1\%}                                       \\
VGG19+VGG16-places365    &                          & 32.22\%          & −2.96                                                & +2.22\%                                             & 29\%          & −13\%                                                & −3\%                                                \\
ResNet50+VGG16-places365 &                          & 28.88\%          & \textbf{3.33}                                        & −1.12\%                                             & 25\%          & −30\%                                                & −7\%                                                \\ \hline
VGG16+VGG16-places365    & \multirow{3}{*}{flatten} & 28.88\%          & −2.97                                                & +0.37\%                                             & 42\%          & −13\%                                                & −10\%                                               \\
VGG19+VGG16-places365    &                          & 28.14\%          & −2.97                                                & −0.37\%                                             & 38\%          & \textbf{−12\%}                                       & −14\%                                               \\
ResNet50+VGG16-places365 &                          & 28.88\%          & −1.12                                                & +0.37\%                                             & 26\%          & −18\%                                                & −26\%                                               \\ \hline
VGG16+VGG16-places365    & \multirow{3}{*}{max}     & 28.51\%          & 0.37                                                 & +1.85\%                                             & 41\%          & −23\%                                                & −10\%                                               \\
VGG19+VGG16-places365    &                          & 27.40\%          & −2.22                                                & +0.74\%                                             & \textbf{44\%} & −17\%                                                & −7\%                                                \\
ResNet50+VGG16-places365 &                          & 28.51\%          & 2.96                                                 & +1.85\%                                             & 33\%          & −28\%                                                & −18\%                                               \\ \hline
\end{tabular}
\label{table5}
\end{adjustwidth}
\end{table}

\section*{Conclusions}

This paper addresses the problem of recognising abuses of human rights given a single image. We present the HRA database, a dataset of images in non-controlled environments containing activities which reveal a human right being violated without any other prior knowledge. The images derive from experts-verified repositories and are labelled with 8 violations categories, proposed and described in this work. Using this dataset and a two-phase deep transfer learning scheme, we conduct an evaluation of recent deep learning algorithms and provide a benchmark on the proposed problem of visual human rights violations recognition. We also presented a thorough investigation on the relevance of combining object-centric and scene-centric CNN features alongside their differences and complementaries. 
A technology capable of identifying potential human rights abuses in the
same way as humans do has a lot of potential applications in human-assistive technologies and would greatly support human rights
investigators.

\section*{Acknowledgments}
This work was supported by the Economic and Social Research Council (ESRC) [grant number ES/ M010236/1] and Engineering and Physical Sciences Research Council (EPSRC) [grant number EP/R02572X/1 and EP/P017487/1].

\section*{Author Contributions}
\textbf{Conceptualization:} GK SE AL KMM\hfill \break
\textbf{Data curation:} GK SE \hfill \break
\textbf{Formal analysis:} GK SE KMM \hfill \break
\textbf{Investigation:} GK SE AL KMM \hfill \break
\textbf{Resources:} GK KMM \hfill \break
\textbf{Software:} GK \hfill \break
\textbf{Validation:} GK SE KMM \hfill \break
\textbf{Writing – original draft:} GK SE  KMM\hfill \break

\nolinenumbers

%
%
%
%
%
%
%

\bibliographystyle{plos2015} 

\begin{thebibliography}{}

\end{thebibliography}


\begin{thebibliography}{10}

\bibitem{United_Nations}
Publication UN. Training Manual on Human Rights Monitoring; 2001.
\newblock Available from:
  \url{http://www.ohchr.org/EN/PublicationsResources/Pages/MethodologicalMaterials.aspx}.

\bibitem{WITNESS}
Matheson K. Video as Evidence Field Guide;.
\newblock Available from: \url{https://witness.org/}.

\bibitem{McPherson}
McPherson E.
\newblock Advocacy Organizations’ Evaluation of Social Media Information for
  NGO Journalism: The Evidence and Engagement Models.
\newblock American Behavioral Scientist. 2015;59(1):124--148.

\bibitem{griffinHolubPerona}
Griffin G, Holub A, Perona P.
\newblock Caltech-256 Object Category Dataset.
\newblock California Institute of Technology; 2007. 7694.

\bibitem{torralba200880}
Torralba A, Fergus R, Freeman WT.
\newblock 80 million tiny images: A large data set for nonparametric object and
  scene recognition.
\newblock IEEE Transactions on Pattern Analysis and Machine Intelligence.
  2008;30(11):1958--1970.

\bibitem{fei2007learning}
Fei-Fei L, Fergus R, Perona P.
\newblock Learning generative visual models from few training examples: An
  incremental bayesian approach tested on 101 object categories.
\newblock Computer Vision and Image Understanding. 2007;106(1):59--70.

\bibitem{zhou2014learning}
Zhou B, Lapedriza A, Xiao J, Torralba A, Oliva A.
\newblock Learning deep features for scene recognition using places database.
\newblock In: Advances in neural information processing systems; 2014. p.
  487--495.

\bibitem{xiao2010sun}
Xiao J, Hays J, Ehinger KA, Oliva A, Torralba A.
\newblock Sun database: Large-scale scene recognition from abbey to zoo.
\newblock In: Computer Vision and Pattern Recognition (CVPR), 2010 IEEE
  conference on. IEEE; 2010. p. 3485--3492.

\bibitem{zhou2016places}
Zhou B, Khosla A, Lapedriza A, Torralba A, Oliva A.
\newblock Places: An image database for deep scene understanding.
\newblock arXiv preprint arXiv:161002055. 2016;.

\bibitem{liu2010exploring}
Liu C, Sharan L, Adelson EH, Rosenholtz R.
\newblock Exploring features in a bayesian framework for material recognition.
\newblock In: Computer Vision and Pattern Recognition (CVPR), 2010 IEEE
  Conference on. IEEE; 2010. p. 239--246.

\bibitem{sharan2009material}
Sharan L, Rosenholtz R, Adelson E.
\newblock Material perception: What can you see in a brief glance?
\newblock Journal of Vision. 2009;9(8):784--784.

\bibitem{bell2015material}
Bell S, Upchurch P, Snavely N, Bala K.
\newblock Material recognition in the wild with the materials in context
  database.
\newblock In: Proceedings of the IEEE conference on Computer Vision and Pattern
  Recognition; 2015. p. 3479--3487.

\bibitem{kalliatakisvisapp17}
Kalliatakis G, Ehsan S, Fasli M, Leonardis A, Gall J, McDonald-Maier KD.
\newblock Detection of Human Rights Violations in Images: Can Convolutional
  Neural Networks Help?
\newblock In: Proceedings of the 12th International Joint Conference on
  Computer Vision, Imaging and Computer Graphics Theory and Applications -
  Volume 5: VISAPP, (VISIGRAPP 2017). INSTICC. SciTePress; 2017. p. 289--296.

\bibitem{kalliatakisHRPDA}
Kalliatakis G, Ehsan S, McDonald-Maier KD.
\newblock A Paradigm Shift: Detecting Human Rights Violations Through Web
  Images.
\newblock In: Proceedings of the Human Rights Practice in the Digital Age
  Workshop; 2017.

\bibitem{t-SNE}
van~der Maaten L, Hinton GE.
\newblock Visualizing High-Dimensional Data Using t-SNE.
\newblock Journal of Machine Learning Research. 2008;9:2579--2605.

\bibitem{VGG}
Simonyan K, Zisserman A.
\newblock Very Deep Convolutional Networks for Large-Scale Image Recognition.
\newblock CoRR. 2014;abs/1409.1556.

\bibitem{caffe}
Jia Y, Shelhamer E, Donahue J, Karayev S, Long J, Girshick R, et~al.
\newblock Caffe: Convolutional Architecture for Fast Feature Embedding.
\newblock In: Proceedings of the 22Nd ACM International Conference on
  Multimedia. MM '14. ACM; 2014. p. 675--678.

\bibitem{ResNet}
He K, Zhang X, Ren S, Sun J.
\newblock Deep Residual Learning for Image Recognition.
\newblock In: 2016 IEEE Conference on Computer Vision and Pattern Recognition
  (CVPR); 2016. p. 770--778.

\bibitem{zhou2017places}
Zhou B, Lapedriza A, Khosla A, Oliva A, Torralba A.
\newblock Places: A 10 million Image Database for Scene Recognition.
\newblock IEEE Transactions on Pattern Analysis and Machine Intelligence.
  2017;.

\bibitem{chollet2015keras}
Chollet F, et~al.. Keras; 2015.
\newblock \url{https://github.com/fchollet/keras}.

\bibitem{transfer_learning}
Pan SJ, Yang Q.
\newblock A Survey on Transfer Learning.
\newblock IEEE Transactions on Knowledge and Data Engineering.
  2010;22(10):1345--1359.

\bibitem{DeCAF}
Donahue J, Jia Y, Vinyals O, Hoffman J, Zhang N, Tzeng E, et~al.
\newblock DeCAF: A Deep Convolutional Activation Feature for Generic Visual
  Recognition.
\newblock In: ICML; 2014. p. 647--655.

\bibitem{Zeiler}
Zeiler MD, Fergus R.
\newblock Visualizing and understanding convolutional networks.
\newblock In: European Conference on Computer Vision. Springer; 2014. p.
  818--833.

\bibitem{kalliatakis_materials}
Kalliatakis G, Stamatiadis G, Ehsan S, Leonardis A, Gall J, Sticlaru A, et~al.
\newblock Evaluating Deep Convolutional Neural Networks for Material
  Classification.
\newblock In: Proceedings of the 12th International Joint Conference on
  Computer Vision, Imaging and Computer Graphics Theory and Applications -
  Volume 5: VISAPP, (VISIGRAPP 2017). INSTICC. SciTePress; 2017. p. 346--352.

\bibitem{oquab2014learning}
Oquab M, Bottou L, Laptev I, Sivic J.
\newblock Learning and transferring mid-level image representations using
  convolutional neural networks.
\newblock In: Proceedings of the IEEE conference on computer vision and pattern
  recognition; 2014. p. 1717--1724.

\bibitem{Krizhevsky}
Krizhevsky A, Sutskever I, Hinton GE.
\newblock ImageNet Classification with Deep Convolutional Neural Networks.
\newblock In: Proceedings of the 25th International Conference on Neural
  Information Processing Systems. NIPS'12. Curran Associates Inc.; 2012. p.
  1097--1105.

\bibitem{zhou2016learning}
Zhou B, Khosla A, Lapedriza A, Oliva A, Torralba A.
\newblock Learning deep features for discriminative localization.
\newblock In: Proceedings of the IEEE Conference on Computer Vision and Pattern
  Recognition; 2016. p. 2921--2929.

\bibitem{selvaraju2016grad}
Selvaraju RR, Das A, Vedantam R, Cogswell M, Parikh D, Batra D.
\newblock Grad-cam: Why did you say that? visual explanations from deep
  networks via gradient-based localization.
\newblock arXiv preprint arXiv:161002391. 2016;.

\bibitem{abadi2016tensorflow}
Abadi M, Agarwal A, Barham P, Brevdo E, Chen Z, Citro C, et~al.
\newblock Tensorflow: Large-scale machine learning on heterogeneous distributed
  systems.
\newblock arXiv preprint arXiv:160304467. 2016;.

\bibitem{mangai2010survey}
Mangai UG, Samanta S, Das S, Chowdhury PR.
\newblock A survey of decision fusion and feature fusion strategies for pattern
  classification.
\newblock IETE Technical review. 2010;27(4):293--307.

\bibitem{snoek2005early}
Snoek CG, Worring M, Smeulders AW.
\newblock Early versus late fusion in semantic video analysis.
\newblock In: Proceedings of the 13th annual ACM international conference on
  Multimedia. ACM; 2005. p. 399--402.

\bibitem{ergun2016early}
Ergun H, Akyuz YC, Sert M, Liu J.
\newblock Early and Late Level Fusion of Deep Convolutional Neural Networks for
  Visual Concept Recognition.
\newblock International Journal of Semantic Computing. 2016;10(03):379--397.

\end{thebibliography}

\end{document}